\documentclass{article}
\usepackage{spconf,amsmath,graphicx}
\usepackage{extra_styles}
\usepackage{multirow}
\usepackage{booktabs}
\usepackage{tabularray}
\usepackage{array}
\usepackage[pagebackref,breaklinks,colorlinks]{hyperref}
\usepackage[subtle]{savetrees}
\usepackage{cite}

\title{SlowFast Network for Continuous Sign Language Recognition}
\name{Junseok Ahn$^*$, Youngjoon Jang$^*$, Joon Son Chung\thanks{$^*$ These authors contributed equally.}}
\address{Korea Advanced Institute of Science and Technology, South Korea}

\begin{document}
\ninept
\maketitle
\begin{abstract}
The objective of this work is the effective extraction of spatial and dynamic features for Continuous Sign Language Recognition (CSLR).
To accomplish this, we utilise a two-pathway SlowFast network, where each pathway operates at distinct temporal resolutions to separately capture spatial (hand shapes, facial expressions) and dynamic (movements) information.
In addition, we introduce two distinct feature fusion methods, carefully designed for the characteristics of CSLR: (1) Bi-directional Feature Fusion (BFF), which facilitates the transfer of dynamic semantics into spatial semantics and vice versa; and (2) Pathway Feature Enhancement (PFE), which enriches dynamic and spatial representations through auxiliary subnetworks, while avoiding the need for extra inference time.
As a result, our model further strengthens spatial and dynamic representations in parallel.
We demonstrate that the proposed framework outperforms the current state-of-the-art performance on popular CSLR datasets, including PHOENIX14, PHOENIX14-T, and CSL-Daily.
\end{abstract}
\begin{keywords}
Continuous sign language recognition, SlowFast network, temporal modelling, bi-directional fusion
\end{keywords}

\section{Introduction}
Sign language serves a pivotal role as a mode of communication for the deaf community in their daily interactions. 
However, due to the challenging nature of sign languages, there are obstacles to direct communication between hard-of-hearing and normal hearing people.
In response to this issue, the field of automatic sign language recognition has received increasing attention with the advances in deep learning techniques. 
Automatic sign language recognition can be divided into two strands: Isolated Sign Language Recognition (ISLR) and Continuous Sign Language Recognition (CSLR).
The former aims to categorise isolated video segments into distinct glosses\footnote{The smallest units with independent meaning in sign language.}. 
On the other hand, the latter progressively transforms video frames into a sequence of glosses, hence being more applicable to real-world scenarios.

To understand continuous sign language and recognise it as meaningful words, it is important to consider spatially meaningful gestures, such as hand direction, finger shape, mouth shape, and facial expression, as well as dynamically meaningful gestures, including arm movements and body gesture. 
For example, sign language's alphabet can be depicted through distinct hand shapes in a single frame. 
Here, the crucial aspect lies in identifying \emph{spatially} meaningful attributes to translate these images into alphabet letters. 
On the other hand, certain words can be portrayed through changes in hand positions over multiple frames. In this instance, the essential task involves identifying \emph{dynamically} significant attributes to interpret the video into the corresponding word.
However, the prevailing CSLR methods~\cite{koller2017re,niu2020stochastic,cheng2020fully, cui2019deep, zhou2020spatial,zuo2022c2slr} predominantly extract features on a frame-by-frame basis, which constrains their ability to capture dynamically significant actions.
Specifically, these methods commonly employ shared 2D Convolutional Neural Networks (CNNs) to capture spatial features. 
This approach results in independent processing of frames without considering interactions with neighbouring frames, consequently limiting their capacity to identify and capitalise on awareness of local temporal patterns that convey sign meanings.
To overcome the inherent limitation of 2D CNNs, some recent studies~\cite{pu2018dilated,pu2019iterative,zhou2019dynamic,ham2021ksl} adopt 3D CNNs~\cite{tran2015learning,carreira2017quo} as their backbone networks. 
However, these approaches still encounter a notable limitation in consolidating spatially distinct information. 
This stems from the fact that the extracted features primarily emphasise only the most crucial regions within spatio-temporal receptive fields. 
Most recent works have aimed to expand both the spatial and temporal receptive fields by computing correlation maps between adjacent frames~\cite{hu2023continuous}. 
However, calculating correlation maps comes with a significant computational burden, posing difficulties in applying this module to high-resolution features.

\begin{figure*}[!t]
   \centering
    \includegraphics[width=0.88\linewidth]{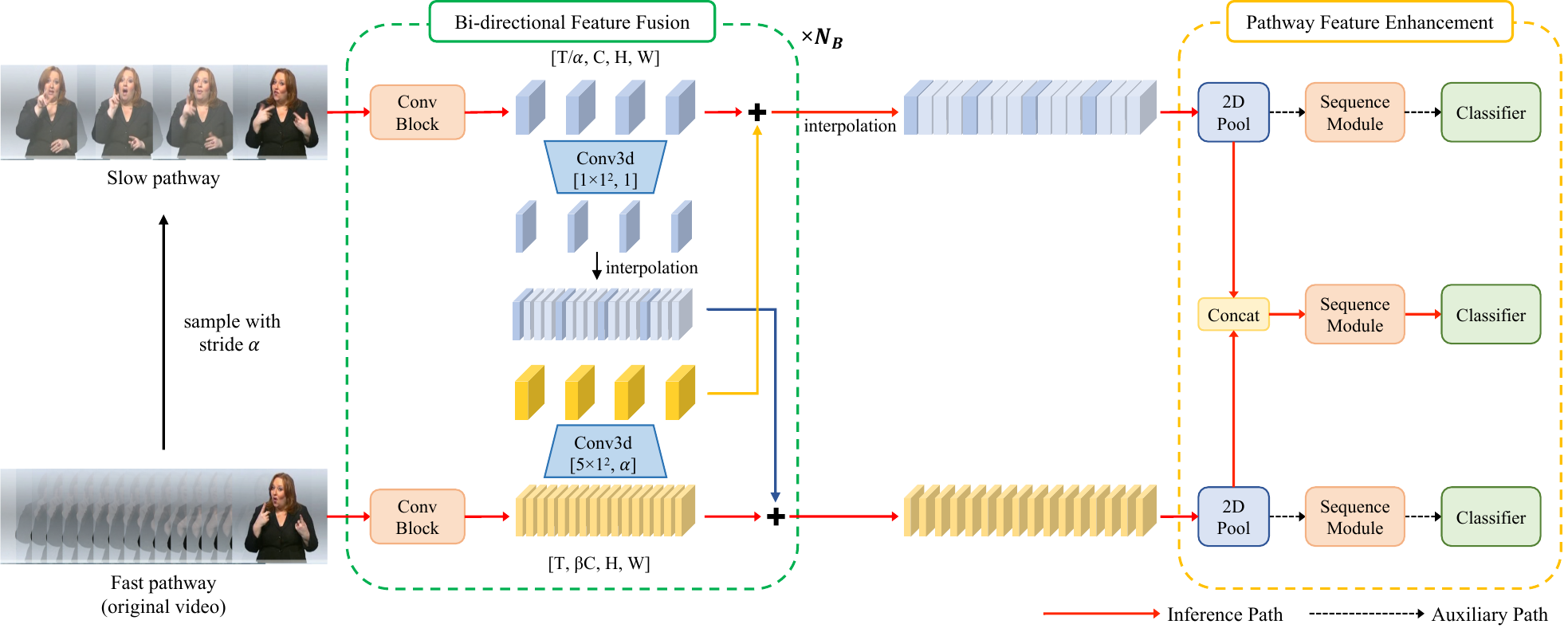}
    \vspace{-3mm}
   \caption{\textbf{Overall Architecture.} Our framework extracts spatial and dynamic features in parallel using the Slow and Fast pathways. Bi-directional Feature Fusion (BFF) facilitates the exchange of rich information between two pathways. Pathway Feature Enhancement (PFE) amplifies the sequence feature of each pathway by utilising characteristics sourced from different sample rates. $\alpha$ and $\beta$ represent the temporal sampling stride and channel reduction ratio, respectively, while $N_B$ stands for the number of blocks in the backbone network.}
    \vspace{-3mm}
   \label{fig:overall}
\end{figure*}

To tackle these aforementioned problems, we adopt the architecture of the SlowFast network~\cite{feichtenhofer2019slowfast}, which is composed of two distinct pathways: (1) Slow pathway, operating at low frame rate to capture spatial semantics, and (2) Fast pathway, operating at high frame rate to capture motion with finer temporal resolution.
The most notable difference between previous literature is that we intentionally extract spatial and dynamic features separately by using two independent subnetworks. 
This is achieved by employing distinct models specialised in capturing either spatial or dynamic aspects.
Given the nature of CSLR where single frame or sequences represent meaningful characters, we introduce the Bi-directional Feature Fusion (BFF) method. This approach efficiently transfers important features to both Slow and Fast pathways with only a minimal increase in computational overhead and model parameters.
Furthermore, since the contents of the Slow and Fast pathways have different characteristics, we also propose Pathway Feature Enhancement (PFE). 
This method enhances sequence features from each pathway, enabling a comprehensive understanding of sign language's spatial and dynamic context.
We evaluate our method on three widely used datasets, PHOENIX14~\cite{koller2015continuous}, PHOENIX14-T~\cite{cihan2018neural}, and CSL-Daily~\cite{zhou2021improving}. Through comprehensive ablation experiments for CSLR, we showcase the effectiveness of the proposed feature fusion methods. Our model establishes state-of-the-art performances across all datasets with notable improvement over the previous works.
\section{Method} 
CSLR and action recognition are distinct tasks. 
Action recognition categorises short video clips into predefined action classes, whereas CSLR is a more intricate task focused on mapping an entire video into a continuous sequence of glosses.
In particular, CSLR requires simultaneous capture of detailed spatial features such as hand shape and facial expressions, as well as dynamic features such as arm movements and body gestures.
This highlights the importance of employing a distinct feature extraction process for different types of actions that convey meaningful signs.

In this paper, we introduce a novel CSLR framework that enables the extraction of spatial and dynamic features in parallel, based on the analysis of CSLR characteristics. 
As illustrated in~\Fref{fig:overall}, our framework comprises (1) the Slow and Fast pathways, (2) Bi-directional Feature Fusion (BFF), and (3) Pathway Feature Enhancement (PFE).
Each pathway serves as an expert in extracting either spatial or dynamic features from sign videos. 
The BFF module facilitates the exchange of complementary information between pathways through a simple adding operation. 
Finally, the PFE fuses sequence features obtained from different frame rates.

\subsection{Slow Pathway}
The Slow pathway extracts spatially important features in sign languages, such as hand shape and facial expressions, from the video data while using a larger temporal stride on input frames. 
The reason for adopting a larger temporal stride lies in the fact that CSLR datasets frequently consist of videos with long durations, necessitating a more efficient approach to processing them. 
An illustrative instance is that the PHOENIX14-T dataset exhibits a maximum frame length of approximately 500 frames.
Nevertheless, using a large stride is not always the optimal choice. 
For instance, action recognition often relies on a large temporal stride of 8, but this can lead to the issue of overlooking rapid and continuous movements, which are essential for conveying meaning in sign language.
Hence, the primary key solution lies in identifying an optimal temporal stride that is both adequately dense to capture essential movements and capable of reducing computational costs. 
Through an ablation study, we select a temporal stride of $\alpha=4$, striking a balance that ensures both performance and computational efficiency.

\subsection{Fast Pathway}
The Fast pathway extracts temporally important dynamics, such as arm movements and body gestures, from the video data with a smaller stride. 
It clearly denotes that the two pathways operate at distinct temporal speeds, resulting in the specialisation of two subnetworks in different types of feature extraction. 
As sign language relies on rapid and intricate movements to convey meaning, it is essential to track these movements. To make sure our model captures all these crucial motions, we set the sampling interval in our Fast pathway to 1.
This indicates that the frame rate of our Fast pathway is four times denser than that of the Slow pathway.
However, as we mentioned earlier, processing all frames of sign video is heavy. 
In response to this, following~\cite{feichtenhofer2019slowfast}, we use significantly lower channel capacity in the Fast pathway, amounting to only $\beta=1/8$ of the Slow pathway's channel size, C. 
This approach results in a more lightweight model, allowing us to increase the depth of the model effectively. 
Furthermore, the limited channel capacity encourages the Fast pathway to focus on temporal modelling by attenuating the ability to represent spatial features.

\subsection{Bi-directional Feature Fusion}
By dividing our model into two subnetworks, we specialise each network for its own purpose. However, to better understand comprehensive signs, it is necessary for each pathway to be aware of the representation learned by the other pathway. To accomplish this objective, we introduce the Bi-directional Feature Fusion (BFF) module, which facilitates the transfer of information between each pathway. 

\newpara{Fast to Slow.}
In the original SlowFast architecture, information from the Fast pathway is integrated into the Slow pathway by concatenating their respective features. 
However, this approach requires the allocation of larger channel sizes for the subsequent convolutional layers. 
To address this limitation, we devise a Fast to Slow pathway fusion mechanism employing addition, instead of concatenation. 
At first, we project the Fast pathway's features by $\alpha$-strided 3D convolution of a $5\times1^2$ kernel with $C$ output channel size. 
This mapping procedure allows the model to learn which information from the Fast pathway is most beneficial for enhancing the Slow pathway's features. 
Finally, we introduce a learnable parameter to weight the importance of the projected features.
Note that this parameter is initialised to 0, allowing the fusion to initially perform an identity mapping and then gradually adapt during training.

\newpara{Slow to Fast.}
We introduce the Slow to Fast fusion pathway, which is not included in the SlowFast baseline~\cite{feichtenhofer2019slowfast}.
The primary objective of this process is to propagate a guide indicating visually important locations on a screen to the Fast pathway. This guide facilitates the extraction of temporal dynamics that constitute meaningful sign motions.
Similar to the Fast to Slow pathway fusion, we project the Slow pathway's features by 1-strided 3D convolution of a $1\times1^2$ kernel with $\beta C$ output channel size. 
Here, we employ a straightforward temporal interpolation to extend the temporal dimension to match the size of the Fast pathway’s features.
The interpolated features are added to the Fast features to transfer spatial information that assists the Fast pathway in capturing both spatially and dynamically meaningful features. 
Note that we use a learnable parameter and initialise it to 0 with the same purpose as the Fast to Slow fusion.

\subsection{Pathway Feature Enhancement}
Unlike action recognition, which maps an output of the backbone network to a single action class using a fully-connected layer, CSLR tasks require the model to predict the gloss sequence in a weakly-supervised manner without access to frame-level gloss labels. For this reason, most recent CSLR models integrate a sequence module on top of the feature extractor. This module aims to identify potential alignments between the visual features and their corresponding labels. Subsequently, these models are optimised by Connectionist Temporal Classification (CTC) loss~\cite{huang2016connectionist}.

\newpara{Feature enhancement.}
Given that the Slow pathway's output possesses a smaller temporal resolution, we perform interpolation to align it with the same temporal resolution as that of the Fast pathway's output. 
The outputs from both pathways are concatenated and subsequently used as input to the main sequence module, as depicted in~\Fref{fig:overall}. 
To delve deeper into the capabilities of the dual pathway encoder structure, we introduce a novel Pathway Feature Enhancement (PFE) method. 
We design two auxiliary sequence modules with the same architecture as the main sequence module. 
This design choice offers two significant advantages: (1) it boosts the representation capacity of each encoder, which aids in temporal alignment by propagating CTC loss, and (2) it ensures that the inference speed is not compromised.

\subsection{Training Details.}
We employ a state-of-the-art architecture for sequence module that includes a 1D CNN followed by two BiLSTM layers. The 1D CNN component is configured as a sequence of $\{K5, P2, K5, P2\}$ layers, where  $K\sigma$ and $P\sigma$ denote a 1D convolutional layer and a pooling layer with kernel size of $\sigma$, respectively. 
Since the output channel sizes of the Slow and Fast pathway encoders are 2048 and 256, respectively, the input channel size for the main sequence module is 2304. 
Both the main and auxiliary sequence modules have a hidden size of 1024.
All pathways are basically trained with CTC loss ($\mathcal{L}_{CTC}$). In addition, following VAC~\cite{min2021visual}, we incorporate the Visual Alignment loss ($\mathcal{L}_{VA}$) and Visual Enhancement loss ($\mathcal{L}_{VE}$) for visual supervision. We formulate the overall loss as follows:
\begin{align}
\mathcal{L}_{main} &= \mathcal{L}_{CTC} + \lambda_{VA}\mathcal{L}_{VA} + \lambda_{VE}\mathcal{L}_{VE}, \\
\mathcal{L}_{slow} &= \mathcal{L}_{CTCs} + \lambda_{VA}\mathcal{L}_{VA_s} + \lambda_{VE}\mathcal{L}_{VE_s}, \\
\mathcal{L}_{fast} &= \mathcal{L}_{CTC_f} + \lambda_{VA}\mathcal{L}_{VA_f} + \lambda_{VE}\mathcal{L}_{VE_f}, \\
\mathcal{L}_{total} &= \mathcal{L}_{main} + \lambda_{slow} \mathcal{L}_{slow} + \lambda_{fast} \mathcal{L}_{fast}.
\end{align}
Here, $\lambda_{slow}$ and $\lambda_{fast}$ represent the loss weights for the two auxiliary losses. 
$\lambda_{VA}$ and $\lambda_{VE}$ are set to 25 and 1, respectively, following VAC methodology.

\section{Experiments}
\subsection{Experimental Setup}
\newpara{Datasets and evaluation metric.} 
The PHOENIX14 dataset contains German weather forecast recordings by 9 actors, comprising 6,841 sentences with 1,295 signs, split into 5,672 training, 540 development, and 629 testing samples. The PHOENIX14-T dataset, designed for CSLR and sign language translation tasks, consists of 8,247 sentences with 1,085 signs, divided into 7,096 training, 519 development, and 642 test instances. The CSL-Daily dataset covers daily life activities by 10 signers, featuring 20,654 sentences distributed among 18,401 training, 1,077 development, and 1,176 test samples.
For evaluation, we adopt Word Error Rate (WER)\footnote{WER = ({\#substitutions} + \text{\#deletions} + \text{\#insertions}) / ({\text{\#words in reference}})}~\cite{koller2015continuous} for evaluation.
\newcolumntype{P}[1]{>{\centering\arraybackslash}p{#1}}
\begin{table}[!t]
    \centering
    \resizebox{0.98\linewidth}{!}{
        \footnotesize
        \begin{tabular}{p{4.3cm}P{0.7cm}P{0.7cm}P{0.7cm}P{0.7cm}}
        \toprule
        \multirow{3}{*}{Methods} & \multicolumn{4}{c}{WER (\%) $\downarrow$} \\
        & \multicolumn{2}{c}{PHOENIX14} & \multicolumn{2}{c}{PHOENIX14-T}\\                 
        & Dev & Test & Dev & Test \\
        \midrule
        \multicolumn{5}{c}{\textbf{RGB Only}} \vspace{3pt} \\
        SFL~\cite{niu2020stochastic}     & 26.2 & 26.8 & 25.1 & 26.1 \\
        FCN~\cite{cheng2020fully}        & 23.7 & 23.9 & 23.3 & 25.1 \\
        Joint-SLRT~\cite{camgoz2020sign} & -    & -    & 24.6 & 24.5 \\
        VAC~\cite{min2021visual}         & 21.2 & 22.3 & -    & -    \\
        LCSA~\cite{zuo2022local}         & 21.4 & 21.9 & -    & -    \\
        CMA~\cite{pu2020boosting}        & 21.3 & 21.9 & -    & -    \\
        SignBT~\cite{zhou2021improving}  & -    & -    & 22.7 & 23.9 \\
        MMTLB~\cite{chen2022simple}      & -    & -    & 21.9 & 22.5 \\
        SMKD~\cite{hao2021self}          & 20.8 & 21.0 & 20.8 & 22.4 \\
        SEN~\cite{hu2023self}        & 19.5 & 21.0 & 19.3 & 20.7 \\
        SSSLR~\cite{jang2023self}        & 20.9 & 20.8 & 20.5 & 22.3 \\
        TLP~\cite{hu2022temporal}        & 19.7 & 20.8 & 19.4 & 21.2 \\
        Corrnet~\cite{hu2023continuous}  & 18.8 & 19.4 & 18.9 & 20.5 \\
        \textbf{SlowFastSign (Ours)}       & \textbf{18.0}     & \textbf{18.3} & \textbf{17.7} & \textbf{18.7} \\
        \midrule
        \multicolumn{5}{c}{\textbf{Additional Modalities}} \vspace{3pt} \\
        DNF (RGB+Optical Flow)~\cite{cui2019deep}                   & 23.1 & 22.9 & -    & -    \\
        SOS (RGB+Scene)~\cite{jang2022signing}              & 20.6 & 21.5 & -    & -    \\
        STMC (RGB+Keypoints)~\cite{zhou2020spatial}              & 21.1 & 20.7 & 19.6 & 21.0 \\
        C$^{2}$SLR (RGB+Keypoints)~\cite{zuo2022c2slr}           & 20.5 & 20.4 & 20.2 & 20.4 \\
        TwoStream-SLR (RGB+Keypoints)~\cite{chen2022two}     & 18.4 & 18.8 & 17.7 & 19.3 \\
        \toprule
        \end{tabular}
    }
    \vspace{-3mm}
    \caption{\textbf{Comparison with state-of-the-art methods on PHOENIX14 and PHOENIX14-T.} Our model achieves new state-of-the-art performance on both datasets.}
    \label{tab:main}
    \vspace{-3mm}
\end{table}

\newpara{Implementation details.}
We adopt SlowFast101~\cite{feichtenhofer2019slowfast} as our baseline with Kinetics-600~\cite{carreira2018short} pretrained weights.
All the input frames are initially resized to $256 \times 256$.
To augment training data, we perform random cropping to achieve a final resolution of $224 \times 224$. We apply horizontal flipping with a probability of 50\% and 20\% temporal rescaling. During inference, we only apply center cropping with a size of $224 \times 224$. Our model is optimised by Adam~\cite{kingma2014adam} optimiser with a weight decay of $10^{-4}$ and a batch size of 2. Our model is trained on the PHOENIX datasets for 80 epochs, and on the CSL-Daily dataset for 50 epochs.
The initial learning rate is $10^{-4}$ for the PHOENIX datasets and reduced by 5 at the 40th and 60th epochs.
For the CSL-Daily dataset, we initialise the learning rate with $5 \times 10^{-5}$ and is halved every 5 epochs starting from 25th epoch.
We empirically set $\lambda_{slow}$ and $\lambda_{fast}$ to 0.25 and 0.25 for PHOENIX datasets and 0.1 and 0.4 for CSL-Daily dataset.

\subsection{Experimental Results}
\newpara{PHOENIX14 and PHOENIX14-T.}
We compare the proposed framework, named SlowFastSign, with previous methods on the PHOENIX14 and PHOENIX14-T datasets. As shown in~\Tref{tab:main}, our model achieves new state-of-the-art performance on both datasets, surpassing the previous best RGB-based method, Corrnet~\cite{hu2023continuous}, by a margin of 1.1\% and 1.8\% on PHOENIX14 and PHOENIX14-T test splits, respectively. We also compare our method with recent works that require additional modalities, such as keypoints and optical flow, for training. Our model outperforms every multimodal-based CSLR model. Specifically, our model exhibits better performance than TwoStream-SLR~\cite{chen2022two} network, which leverages human keypoints. The margins of improvement are 0.5\% on PHOENIX14 and 0.6\% on PHOENIX14-T, respectively. We highlight that the proposed method does not require additional modalities or annotations to identify spatially important regions.

\newpara{CSL-Daily.}
We also compare our framework on the CSL-Daily dataset to demonstrate its generalisability. As reported in~\Tref{tab:csl}, the proposed framework outperforms all the previous state-of-the-art methods. Notably, our method shows a 5.2\% improvement compared to the RGB-based Corrnet. Furthermore, our model excels the TwoStream-SLR network, which incorporates additional supervision, by a 0.4\% margin on test split.

\begin{table}[!t]
    \centering
    \resizebox{0.92\linewidth}{!}{
        \footnotesize
        \begin{tabular}{p{5cm}P{0.9cm}P{0.9cm}}
        \toprule
        \multirow{2}{*}{Methods} & \multicolumn{2}{c}{WER (\%) $\downarrow$} \\             
        & Dev & Test \\
        \midrule
        \multicolumn{3}{c}{\textbf{RGB Only}} \vspace{3pt} \\
        LS-HAN$^\dag$~\cite{huang2018video} & 39.0 & 39.4 \\
        SignBT~\cite{zhou2021improving} & 33.6 & 33.1 \\
        FCN$^\dag$~\cite{cheng2020fully} & 33.2 & 32.5 \\
        DNF (RGB)$^\dag$~\cite{cui2019deep} & 32.8 & 32.4 \\
        Joint-SLRT$^\dag$~\cite{camgoz2020sign} & 33.1 & 32.0 \\
        Corrnet~\cite{hu2023continuous}  & 30.6 & 30.1 \\
        \textbf{SlowFastSign (Ours)}     & 25.5 & \textbf{24.9}  \\
        \midrule
        \multicolumn{3}{c}{\textbf{Additional Modalities}} \vspace{3pt} \\
        TwoStream-SLR (RGB+Keypoints)~\cite{chen2022two} & \textbf{25.4} & 25.3 \\
        \toprule
        \end{tabular}
    }
    \vspace{-1mm}
    \caption{\textbf{Comparison with state-of-the-art methods on CSL-Daily dataset.} $^{\dagger}$ indicates the reproduced results in~\cite{zhou2021improving}. Our model achieves new state-of-the-art performance.}
    \label{tab:csl}
\end{table}

\begin{table}[!t]
    \centering
    \resizebox{0.792\linewidth}{!}{
        \footnotesize
        \begin{tabular}{P{0.2cm}P{0.7cm}P{0.7cm}P{2.2cm}P{1.3cm}}
        \toprule
        \multirow{2}{*}{$\alpha$} & \multicolumn{2}{c}{WER (\%) $\downarrow$} & Maximum & \multirow{2}{*}{GFLOPs (G)}\\
        & Dev & Test & memory usage (MB) & \\
        \midrule
        8 & 20.9 & 21.5 & 13,829 & 283.9 \\
        4 & 18.0 & 18.3 & 21,418 & 521.2 \\
        2 & 17.4 & 17.9 & 39,521 & 995.8 \\
        \toprule
        \end{tabular}
    }
    \vspace{-1mm}
    \caption{\textbf{Effect of temporal stride $\alpha$.} We observe that a denser temporal resolution achieved by lower $\alpha$ in the Slow pathway steadily improves the CSLR performance.}
    \label{tab:stride}
\end{table}

\subsection{Ablation Study}
\newpara{Effect of temporal stride $\alpha$ in Slow pathway.}
In order to investigate the impact of temporal stride $\alpha$ in the Slow pathway, we ablate the value of $\alpha$ in~\Tref{tab:stride}. As we increase the input temporal resolution of the Slow pathway, we observe a consistent improvement in our model's performance. However, this leads to higher memory usage and computational overhead, which hinders real-world applications and deployment of CSLR.
For fair comparisons with previous works considering computational costs and memory usage, we use $\alpha=4$ in all of our experiments.

\newpara{Design choice of Bi-directional Feature Fusion.}
As shown in~\Tref{tab:bifuse}, the BFF module utilising concatenation yields a 0.4\% improvement in test performance compared to the baseline, demonstrating the effectiveness of the bi-directional fusion approach. However, this comes at the cost of an additional 8 GFLOPs to process 200-frame video.
On the other hand, with the addition method, our BFF module not only reduces the computational cost by 3 GFLOPs compared to the previous method but also shows the best performance among the three methods.
Additionally, our model has even fewer parameters than the uni-directional baseline.

\newpara{Component analysis.}
To establish the effectiveness of all components within our framework, we systematically conduct ablation experiments by progressively incorporating the proposed components into the baseline model in~\Tref{tab:ablation}. Initially, we assess the performance of the individual Slow and Fast networks separately. The model using only the Slow pathway exhibits superior recognition performance compared to the model using only the Fast pathway. This emphasises that the extraction of spatial semantics is crucial in CSLR tasks. However, when both Slow and Fast pathways are employed (utilising the same architecture as the original SlowFast network, which includes only Fast-to-Slow feature fusion), it becomes evident that the temporal information extracted from the Fast pathway plays a pivotal role in capturing motion. With the incorporation of the BFF module, it surpasses the performance of the vanilla SlowFast network, affirming the effectiveness of propagating spatial information to the Fast pathway in CSLR. In the end, by integrating all proposed components, we attain the highest score, showing a performance improvement of 0.6\% on the both dev and test sets in comparison to the SlowFast baseline.
\begin{table}[!t]
    \centering
    \resizebox{0.99\linewidth}{!}{
        \footnotesize
        \begin{tabular}{p{2.7cm}P{0.8cm}P{0.8cm}P{1.25cm}P{1.3cm}}
        \toprule
        \multirow{2}{*}{Methods} & \multicolumn{2}{c}{WER (\%) $\downarrow$} & \multirow{2}{*}{\#Params (M)} & \multirow{2}{*}{GFLOPs (G)} \\
        & Dev & Test & \\
        \midrule
        SlowFast & 18.6 & 18.9 & 52.9 & 515.9 \\
        SlowFast + BFF (concat) & 18.6 & 18.5 & 53.1 & 524.0 \\
        SlowFast + BFF (add) & 18.4 & 18.3 & 52.5 & 521.2 \\
        \toprule
        \end{tabular}
    }
    \vspace{-1mm}
    \caption{\textbf{Design choice on Bi-directional Feature Fusion.} We design BFF module with a projector and an adding operation to minimise the computational costs and modal parameters.}
    \label{tab:bifuse}
\end{table}
\begin{table}[t!]
    \centering
    \resizebox{0.85\linewidth}{!}{
        \begin{tabular}{P{1.1cm}P{1.1cm}P{1.1cm}P{1.1cm}P{1cm}P{1cm}}
        \toprule
        & & & & \multicolumn{2}{c}{WER (\%) $\downarrow$} \\
        Slow & Fast & BFF & PFE & Dev & Test \\
        \midrule
        \cmark & & & & 18.9 & 20.0 \\
        & \cmark & & & 24.1 & 24.3 \\
        \cmark & \cmark & & & 18.6 & 18.9 \\
        \cmark & \cmark & \cmark & & 18.4 & 18.3 \\
        \cmark & \cmark & & \cmark & 18.3 & 18.4 \\
        \midrule
        \cmark & \cmark & \cmark & \cmark & \textbf{18.0}     & \textbf{18.3}  \\
        \toprule
        \end{tabular}
    }
    \vspace{-1mm}
    \caption{\textbf{Analysis of the proposed components.} Each component consistently contributes to boosting performance. The best result is achieved when all components are combined.}
    \label{tab:ablation}
\end{table}

\begin{figure}[!t]
    \centering
    \includegraphics[width=1\linewidth]{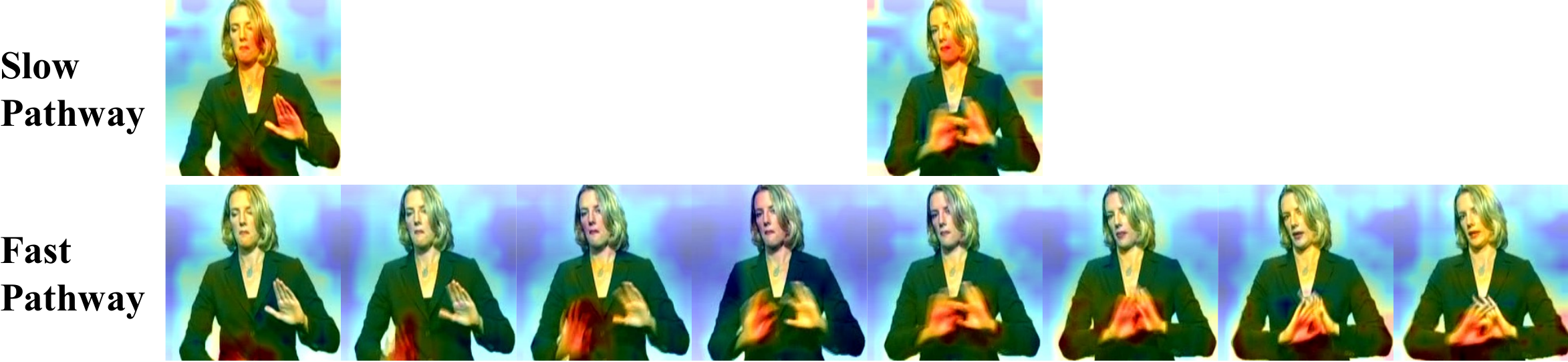}
    \vspace{-0.4mm}
    \caption{\textbf{GradCAM visualisation.} The Slow pathway emphasises hand and face regions, which are crucial components in sign language, while the Fast pathway captures dynamic movements.}
    \label{fig:gradcam}
\end{figure}

\newpara{GradCAM Visualisation.}
To demonstrate that the Slow and Fast pathways capture different types of features, we visualise each pathway's feature maps using GradCAM~\cite{selvaraju2017grad}. As shown in~\Fref{fig:gradcam}, the Slow pathway highlights both hand and face regions, which are essential for sign language. On the other hand, as shown in the figure, the Fast pathway focuses heavily on the dynamically moving hand of the signer. 
This implies that our framework is capable of capturing both spatial and dynamic features in parallel within a screen.

\section{Conclusion}
In this paper, we aim to extract both spatial and dynamic features for Continuous Sign Language Recognition (CSLR). To accomplish this goal, we employ the SlowFast network, which captures spatial and dynamic features with two expert networks operating at different temporal resolutions. To the best of our knowledge, we are the first to separately extract distinct types of features, tailored to the specific characteristics of sign language. Additionally, we introduce two novel methods that improve the capacity of each pathway: (1) Bi-directional Feature Fusion (BFF), which facilitates the exchange of temporal and spatial semantics, and (2) Pathway Feature Enhancement (PFE), which enhances dynamic and spatial representations through auxiliary subnetworks, while avoiding the need for additional inference time. The experimental results show that our framework achieves state-of-the-art performance on three large-scale benchmarks.



\bibliographystyle{IEEEbib}
\bibliography{shortstrings,refs}

\end{document}